\documentclass[10pt,letterpaper]{article}

\usepackage{cogsci}

\cogscifinalcopy 

\usepackage{cmap}
\usepackage[T1]{fontenc}
\usepackage[american]{babel}
\usepackage{csquotes}
\usepackage{newtxtext,newtxmath}  
\usepackage{fontawesome5}
\usepackage{contour}
\usepackage[normalem]{ulem}

\usepackage[
  backend=biber,
  style=apa,
  natbib=true,
  eprint=true,
  annotation=false,
]{biblatex}
\usepackage{float}
\usepackage{graphicx}

\contourlength{0.8pt}

\newcommand{\myuline}[1]{%
  \uline{\phantom{#1}}%
  \llap{\contour{white}{#1}}%
}

\usepackage[hidelinks]{hyperref}

\title{Neurally-plausible radial basis kernels using distributed Fourier embeddings}

\author[1,2]{\mbox{Jakeb Chouinard (jakeb.chouinard@uwaterloo.ca)}}
\affil[1]{Department of Systems Design Engineering, University of Waterloo}
\affil[2]{Centre for Theoretical Neuroscience, University of Waterloo}

\begin{document}

\maketitle

\begin{abstract}
Coherent, continuous spatial representations are critical for synthesizing physical and perceptual phenomena into a single representational space. Radial basis kernels provide a path forward for this type of distributed representation. In this work, we aim to characterize and analyze common radial basis kernels realizable in the neurally-plausible framework of spatial semantic pointers. Further, we analyze previous radial basis kernel work based on grid cell-like representations and demonstrate that such representations are both capable of and optimal for realizing radial basis kernels\footnote{Code implementing the discussed SSP representations is available in the \faGithub\,\,\myuline{\href{https://github.com/ctn-waterloo/sspspace/tree/main}{``sspspace'' Python package}} as kernel configurations of ``JointSSPSpace'' and ``HexagonalSSPSpace'' encoders.}.

\end{abstract}

\section{Introduction}

Cohesive representation of multidimensional spaces is essential for cognition. In spatial cognition, different equidistant points within a space, such as corners of a trapezoidal room when one is at its centroid, should maintain their sense of equidistance when represented in neurons. We can consider their equidistance as a function of kernels approximated by the inner-product similarities of these neurally-plausible distributed representations. Traditional Fourier-based embeddings, such as Random Fourier Features (RFFs; \cite{rahimi_random_2007}) and Spatial Semantic Pointers (SSPs; \cite{komer_neural_2019}), realize single-dimensioned kernels. When multiple of these embeddings are used in tandem, kernels become products of the individual single-dimensioned kernels \citep{voelker_short_2020}. By consequence, the inner-product similarity of equidistant points is dependent on both the direction of the vector between the two points and its magnitude. By contrast, radial basis functions calculate similarity independent of direction. HexSSPs are a form of spatial semantic pointer that exhibit radial basis function-like kernels \citep{dumont_accurate_2020, stockel_assorted_2020}; however, the exact nature of the HexSSP kernel has yet to be characterized in an analytical way. Further, we examine how alternative sampling methods can, using grid cell-like representations, approximate any form of radial basis kernel.

Since there is a direct connection between the distribution that the Fourier-embedding phases of the HexSSPs approximate and the resultant kernel, we aim to first characterize and understand how direction-uniform, or isotropic, distributions can shape kernels. We then analyze the composition of HexSSPs to relate them back to these isotropic distributions, providing an analytical formation for the HexSSP kernel, the hypergeometric kernel, as well as two alternative kernels---the Gaussian and $n$-jinc kernels.

\section{Methods}

Consistent with previous research on kernels and their relationship to Fourier embeddings, we refer to Bochner's theorem to conveniently transition between probability distributions in the Fourier space and their resultant shift-invariant kernels since probability density functions are positive-finite Borel measures \citep{bochner_lectures_1959}. In brief, this can be written as:
\begin{equation}
    K(\boldsymbol{x})=\mathcal{F}^{-1}\left\{p(\boldsymbol{\omega})\right\} \notag
\end{equation}
or
\begin{equation}
    K(\boldsymbol{x})=\int_{\Omega}e^{j\boldsymbol{\omega}^{\top}\boldsymbol{x}}p(\boldsymbol{\omega})d\boldsymbol{\omega} \notag
\end{equation}
where $\mathcal{F}^{-1}\left\{\cdot\right\}$ is the Fourier inverse transform, $\boldsymbol{x},\boldsymbol{\omega}\in\mathbb{R}^{n}$ are feature and phase space vectors respectively, $p(\boldsymbol{\omega})\in\mathbb{R}^{+}$ is the probability density function of $\boldsymbol{\omega}$ over $\Omega$, and $j$ is the imaginary unit. It is worth noting that the opposite case is true as well---$p(\boldsymbol{\omega})$ can be calculated for an arbitrary kernel by calculating its Fourier transform: $p(\boldsymbol{\omega})=\mathcal{F}\left\{K(\boldsymbol{x})\right\}$.

\subsection{A Generalized Approach to Radial Basis Kernels}
In order to realize a radial basis function as a kernel over an $n$-dimensional feature space, we must sample rows of some phase matrix, $A\in\mathbb{R}^{d\times n}$, from a joint probability distribution defined over the $n$-dimensional phase space. Since $A$ is conjugate symmetric to enforce real-valued embeddings, the joint probability distribution will be centred and symmetric when considering all non-zero rows of $A$. We can consider that the resultant probability density function will be defined uniformly in every direction within the $n$-dimensional phase space, making the joint probability distribution of $\boldsymbol{\omega}$ independent in terms of direction and magnitude.

Let $p(\boldsymbol{\omega})$ be the joint probability distribution for sampled rows of $A$. Given that the magnitude and direction are independent, $p(\boldsymbol{\omega})=p(\|\boldsymbol{\omega}\|_{2},\hat{\boldsymbol{\omega}})=p(\|\boldsymbol{\omega}\|_{2})p(\hat{\boldsymbol{\omega}})$ where $\|\boldsymbol{\omega}\|_{2}$ is the $\ell$-2 norm of the phase vector (magnitude) and $\hat{\boldsymbol{\omega}}$ is the unit phase vector (direction). It follows that $p(\boldsymbol{w})$ is an isotropic or rotation-invariant distribution.

Since $\hat{\boldsymbol{\omega}}$ is uniform in every direction within the $n$-dimensional space, the distribution of $\hat{\boldsymbol{\omega}}$ is the uniform distribution of points on the surface of the unit $n$-ball. The uniform directional distribution can be written as the inverse of the surface area of the unit $n$-ball, $\mathcal{A}_{n}$, such that $p(\hat{\boldsymbol{\omega}})=\mathcal{A}_{n}^{-1}$
where $\mathcal{A}_{n}=2\pi^{n/2}\left(\Gamma\left(\frac{n}{2}\right)\right)^{-1}$. Consequently, $p(\boldsymbol{\omega})$ becomes:
\begin{equation}
    p(\boldsymbol{\omega})=p(\|\boldsymbol{\omega}\|_{2})\frac{\Gamma\left(\frac{n}{2}\right)}{2\pi^{n/2}} \notag
\end{equation}
and the kernel approximation becomes:
\begin{equation}
    K(\boldsymbol{x})=\int_{\Omega}e^{j\boldsymbol{\omega}^{\top}\boldsymbol{x}}p(\|\boldsymbol{\omega}\|_{2})\frac{\Gamma\left(\frac{n}{2}\right)}{2\pi^{n/2}}d\boldsymbol{\omega} \notag
\end{equation}

Since the magnitude and direction are independent, we can rearrange the integral into directional and magnitudinal components by splitting the volumetric integral into directional and magnitudinal integrals:
\begin{equation}
    K(\boldsymbol{x})=\frac{\Gamma\left(\frac{n}{2}\right)}{2\pi^{n/2}}\iint_{\|\boldsymbol{\omega}\|_{2},\hat{\boldsymbol{\omega}}}e^{j\boldsymbol{\omega}^{\top}\boldsymbol{x}}p(\|\boldsymbol{\omega}\|_{2})d\|\boldsymbol{\omega}\|_{2}d\hat{\boldsymbol{\omega}} \notag
\end{equation}
where components lacking dependence on direction can be shifted to form and inner and outer integral of direction and magnitude respectively. Let $r=\|\boldsymbol{\omega}\|_{2}$:
\begin{equation}
    K(\boldsymbol{x})=\frac{\Gamma\left(\frac{n}{2}\right)}{2\pi^{n/2}}\int_{r}p(r)\left(\int_{\hat{\boldsymbol{\omega}}}e^{jr\hat{\boldsymbol{\omega}}^{\top}\boldsymbol{x}}d\hat{\boldsymbol{\omega}}\right)dr   \notag 
\end{equation}
where the inner integral is a spherical plane wave integration such that:
\begin{align}
    K(\boldsymbol{x})=\frac{\Gamma\left(\frac{n}{2}\right)}{2\pi^{n/2}}\int_{0}^{\infty}p(r)(2\pi)^{n/2}(r\|\boldsymbol{x}\|_{2})^{\frac{n}{2}-1}J_{\frac{n}{2}-1}(r\|\boldsymbol{x}\|_{2})dr \notag
\end{align}
or:
\begin{align} \label{eq:gen_kernel}
    K(\boldsymbol{x})=\Gamma\left(\frac{n}{2}\right)\int_{0}^{\infty}p(r)\left(\frac{2}{r\|\boldsymbol{x}\|_{2}}\right)^{\frac{n}{2}-1}J_{\frac{n}{2}-1}(r\|\boldsymbol{x}\|_{2})dr
\end{align}

Equation \ref{eq:gen_kernel} provides a general form of the similarity kernel for an $n$-dimensional feature space. As an example, we can consider uniform sampling in a single-dimensioned feature space. For a phase matrix in which samples are uniformly sampled, we would expect $K(x)$ to approximate a sinc function---$\mathrm{sinc}(x)=x^{-1}\sin(x)$. For $r\sim\mathcal{U}(0,1)$ and $n=1$ such that $p(r)=1$ if $r\in[0,1]$ and 0 elsewhere:
\begin{align}
    K(x)&=\sqrt{\pi}\int_{0}^{1}\left(\frac{2}{rx}\right)^{-\frac{1}{2}}J_{-\frac{1}{2}}(rx)dr \notag \\
    &=\sqrt{\pi}\int_{0}^{1}\sqrt{\frac{rx}{2}}\sqrt{\frac{2}{\pi rx}}\cos(rx)dr \notag \\
    &=\int_{0}^{1}\cos(rx)dr \notag \\
    &=\mathrm{sinc}(x) \notag
\end{align}

Alternatively, we may find it desirable to realize a Gaussian kernel for any $n$-dimensional feature space. Let $r\sim\chi(n)$:
\begin{align}
    K(\boldsymbol{x})&=\Gamma\left(\frac{n}{2}\right)\int_{0}^{\infty}\frac{r^{n-1}e^{-\frac{r^{2}}{2}}}{2^{\frac{n}{2}-1}\Gamma\left(\frac{n}{2}\right)}\left(\frac{2}{r\|\boldsymbol{x}\|_{2}}\right)^{\frac{n}{2}-1}J_{\frac{n}{2}-1}(r\|\boldsymbol{x}\|_{2})dr \notag \\
    &=\int_{0}^{\infty}\frac{r^{\frac{n}{2}}e^{-\frac{r^{2}}{2}}}{\|\boldsymbol{x}\|_{2}^{\frac{n}{2}-1}}J_{\frac{n}{2}-1}(r\|\boldsymbol{x}\|_{2})dr \notag \\
    &=e^{-\frac{\|\boldsymbol{x}\|_{2}}{2}}
\end{align}

Equation 2 demonstrates that a Gaussian, symmetric kernel can be realized for any $A\in\mathbb{R}^{d\times n}$ by sampling its rows from an isotropic distribution where $r\sim\chi(n)$.

\subsection{Grid Cells as Radial Basis Kernel Primitives}
Previous work has shown that kernels with hexagonal tiling can be constructed using a centred simplex to generate rows of $A$ \citep{dumont_accurate_2020}. These representations have additionally been implemented in spiking neural models to demonstrate grid cell-like activations in spatial navigation problems, such as Simultaneous Localization and Mapping \citep{dumont_exploiting_2023}. For a single grid cell-like representation, the phase matrix's rows are equal to the points of the centred $n$-simplex multiplied by some scaling term, resulting in $A\in\mathbb{R}^{2(n+1)+1\times n}$. For a place cell-like representation, we can concatenate $N_{R}$ different grid cell orientations and $N_{S}$ different scales of the $n$-simplex into $A$ such that $A\in\mathbb{R}^{2N_{R}N_{S}(n+1)+1\times n}$. Embeddings with an $A$ of this structure are called Hexagonal Spatial Semantic Pointers---or HexSSPs. Comparably little work has been done to analyze the kernels induced by these HexSSPs; however qualitatively, they demonstrate radial symmetry with some amount of oscillatory behaviour for sufficient $N_{R}$ and $N_{S}$. Beyond qualitative analysis, there is limited information available.

To analyze the resultant kernels, we first consider the distribution of the phase vectors' directions. HexSSPs sample rotation transforms from the special orthogonal group in $\mathbb{R}^{n}$---$SO(n)$---using the Haar distribution to create multiple grid cell-like representations of varying orientation. Supposing every simplex vertex was $\ell$-2 distance of $1$ from $\boldsymbol{0}$, the Haar distribution, which is uniform over the rotational space, uniformly rotates the vertices over the surface of the $n$-ball. While each simplex maintains structure between its individual points, the resultant set of points for all simplices approximates a uniform distribution of points over the surface of the $n$-ball. In connection to the previous section, this approximates an isotropic distribution of points as $N_{R}\rightarrow\infty$, so $p(\hat{\boldsymbol{\omega}})=\mathcal{A}^{-1}_{n}$. For some $N_S$, HexSSPs scale the individual simplices by some scalar value. This is the same process described previously where uniform directions could be scaled by some distribution of magnitudes, $p(r)$, to approximate some underlying centred, symmetric joint distribution, $p(\boldsymbol{\omega})$. We can use Equation \ref{eq:gen_kernel} to characterize the kernel approximated by HexSSPs. Traditionally, HexSSPs use a uniform distribution of magnitudes to embed some multidimensional feature space ($n>1$). We begin by considering the $n$-dimensional case where $r\sim\mathcal{U}(0,1)$:
\begin{align}
    K(\boldsymbol{x})=\Gamma\left(\frac{n}{2}\right)\int_{0}^{1}\left(\frac{2}{r\|\boldsymbol{x}\|_{2}}\right)^{\frac{n}{2}-1}J_{\frac{n}{2}-1}(r\|\boldsymbol{x}\|_{2})dr \notag
\end{align}
Let $\nu=\frac{n}{2}-1$:
\begin{align}
    K(\boldsymbol{x})&=\Gamma\left(\nu+1\right)\left(\frac{2}{\|\boldsymbol{x}\|_{2}}\right)^{\nu}\int_{0}^{1}r^{-\nu}J_{\nu}(r\|\boldsymbol{x}\|_{2})dr \notag
\end{align}

Since $J_{\nu}(r\|\boldsymbol{x}\|_{2})$ can be rewritten as a function of a generalized hypergeometric function, $\,_{0}F_{1}$, we substitute and simplify further using the relationship:
$$J_{\nu}(x)=\left(\frac{x}{2}\right)^{\nu}\frac{1}{\Gamma(\nu+1)}\,_{0}F_{1}\left(\,\,\nu+1;\,\,-\frac{x^{2}}{4}\right)$$
and since all other terms cancel out:
\begin{align}
    K(\boldsymbol{x})&=\int_{0}^{1}\,_{0}F_{1}\left(\nu+1;\,-\frac{\left(r\|\boldsymbol{x}\|_{2}\right)^{2}}{4}\right)dr \notag
\end{align}
For any uniform distribution $r\sim\mathcal{U}(0,\lambda)$, the kernel becomes:
\begin{align}
    K(\boldsymbol{x})&=\frac{1}{\lambda}\int_{0}^{\lambda}\,_{0}F_{1}\left(\nu+1;\,-\frac{\left(r\|\boldsymbol{x}\|_{2}\right)^{2}}{4}\right)dr \notag
\end{align}
where an increased $\lambda$ leads to a decrease in the peak width of the kernel. Most SSP literature uses a length-scale parameter, $\ell\in\mathbb{R}^{+}$, to inversely scale $A$ during embedding. This is equivalent to $\lambda=\ell^{-1}$. Perhaps more intuitively, an increased $\ell$ leads to an increased kernel peak width. In pursuit of a more condensed form, we can expand the hypergeometric kernel using the power series definition of the hypergeometric function and integrate each term individually, resulting in:
\begin{equation}
    K(\boldsymbol{x})=\sum_{k=0}^{\infty}\frac{(-1)^{k}\ell^{-2k}\|\boldsymbol{x}\|_{2}^{2k}}{4^{k}(2k+1)(\frac{n}{2})_{k}k!}
\end{equation}
where $(\cdot)_{k}$ is the Pochhammer symbol: $(x)_{k}=\prod_{i=0}^{k-1}(x+i)$.

\begin{figure}[h!]
    \centering
    \includegraphics[width=1.0\linewidth]{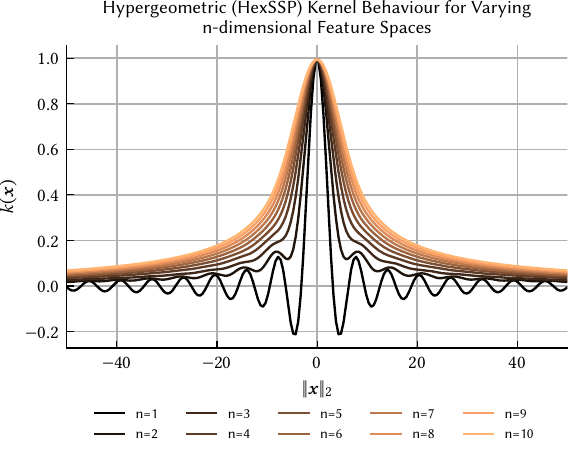}
    \caption{HexSSP hypergeometric kernel shape for varying $n$-dimensional feature spaces from Equation 3 ($\lambda=\ell^{-1}=1$).}
    \label{fig:hexssp-kernels-uniform-radius}
\end{figure}

 We refer to this resultant kernel---the one most commonly used in HexSSP literature---as the ``integrated hypergeometric kernel'' or the ``hypergeometric kernel'' for brevity. Figure \ref{fig:hexssp-kernels-uniform-radius} shows the hypergeometric kernel for varying values of $n$. We see an evident widening of the kernel as the value of $n$ increases with an accompanying decrease in oscillations, approaching a bell-like curve. While the kernel becomes smoother and widens as the value of $n$ increases, $\lim_{n\rightarrow\infty}K(\boldsymbol{x})=1\,\,\forall\,\,\boldsymbol{x}\in\mathbb{R}^{n}$ since $(\cdot)_{0}=1$. Consequently, for a sufficiently high number of features, the embedding space can no longer distinguish between them. This is unlike the Gaussian kernel derived in Equation 2, which is evidently ambivalent to the dimensionality of the feature space; this is due to the feature space dimensionality dependence for $p(r)$ where $r\sim\chi(n)$. Note that we can adjust the peak-width of the Gaussian kernel using $\ell^{-1}$ just as we did for the hypergeometric kernel. Since $r\sim\chi(n)=\sqrt{\mathrm{\Gamma}\left(\frac{n}{2},2\right)}$ where $\Gamma(\cdot,\cdot)$ is the gamma distribution, scaling $r$ by $\ell^{-1}$ results in $r\sim\sqrt{\mathrm{\Gamma}\left(\frac{n}{2},2\ell^{-2}\right)}$, and the Gaussian kernel becomes:
 \begin{equation}
     K(\boldsymbol{x})=e^{-\frac{\ell^{-2}\|\boldsymbol{x}\|_{2}^{2}}{2}}
 \end{equation}

Lastly, we consider $r\sim\beta_{\lambda}(n,1)$---a beta distribution such that $p(r)=n\lambda^{-n}r^{n-1}$ over its support, $r\in[0,\lambda]$, and 0 elsewhere. The scaled beta distribution ensures that phase vectors are uniformly distributed within an $n$-ball of radius $\lambda$:
\begin{equation}
    K(\boldsymbol{x})=\Gamma\left(\frac{n}{2}\right)\int_{0}^{\lambda}n\lambda^{-n}r^{n-1}\left(\frac{2}{r\|\boldsymbol{x}\|_{2}}\right)^{\frac{n}{2}-1}J_{\frac{n}{2}-1}(r\|\boldsymbol{x}\|_{2})dr \notag
\end{equation}
and with some rearranging:
\begin{equation}
    K(\boldsymbol{x})=\Gamma\left(\frac{n}{2}\right)\left(\frac{2}{\|\boldsymbol{x}\|_{2}}\right)^{\frac{n}{2}-1}n\lambda^{-n}\int_{0}^{\lambda}r^{\frac{n}{2}}J_{\frac{n}{2}-1}(r\|\boldsymbol{x}\|_{2})dr \notag
\end{equation}
where the integral is a simple Bessel identity such that:
\begin{equation}
    K(\boldsymbol{x})=\Gamma\left(\frac{n}{2}\right)\left(\frac{2}{\|\boldsymbol{x}\|_{2}}\right)^{\frac{n}{2}-1}n\lambda^{-n}\frac{\lambda^{\frac{n}2}}{\|\boldsymbol{x}\|_{2}}J_{\frac{n}{2}}(\lambda\|\boldsymbol{x}\|) \notag
\end{equation}
or, rearranging and replacing $\lambda$ with $\ell^{-1}$ for consistency with SSP literature:
\begin{equation}
    K(\boldsymbol{x})=2^{\frac{n}{2}-1}n\Gamma\left(\frac{n}{2}\right)\frac{J_{\frac{n}{2}}(\ell^{-1}\|\boldsymbol{x}\|)}{\left(\ell^{-1}\|\boldsymbol{x}\|_{2}\right)^{\frac{n}{2}}}
\end{equation}

\begin{figure}[h!]
    \centering
    \includegraphics[width=1.0\linewidth]{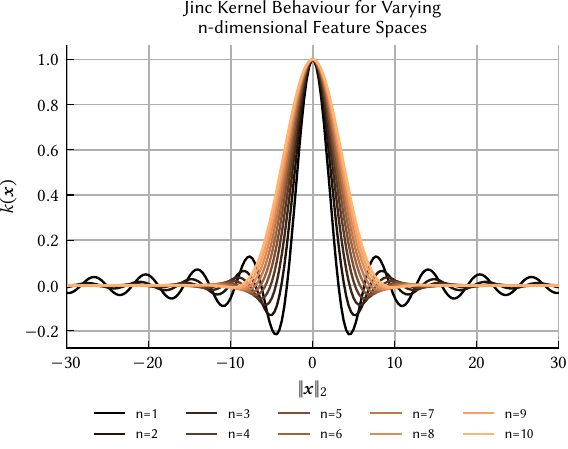}
    \caption{Jinc kernel shape for varying $n$-dimensional feature spaces from Equation 5 ($\lambda=\ell^{-1}=1$).}
    \label{fig:hexssp-kernels-uniform-sphere}
\end{figure}

Equation 5 describes a kernel that is a decaying Bessel function. We refer to this resultant kernel as an ``$n$-jinc kernel'' since it is exactly a jinc function of the magnitude when $n=2$---$\mathrm{jinc}(x)=x^{-1}J_{0}(x)$. Similar to the hypergeometric kernel, the jinc kernel changes in shape for varying $n$ and $\lim_{n\rightarrow\infty}K(\boldsymbol{x})=1\,\forall\,\boldsymbol{x}\in\mathbb{R}^{n}$. We can see this by replacing the Bessel function in Equation 5 with the hypergeometric relationship specified previously and expanding it into a power series representation as we did to evaluate the hypergeometric kernel's integral. Unlike the Gaussian kernel, dependence on $n$ within the distribution of $r$ to ensure a consistent volumetric density distribution does not prevent kernel changes with respect to $n$. Figure \ref{fig:hexssp-kernels-uniform-sphere} shows the jinc kernel for varying $n$-dimensional feature spaces.

Since the HexSSP generation structure is capable of using any sampling method for its magnitudes, we can approximate any symmetric, centred probability distribution, $p(r)$, and its corresponding radial basis kernel. Taking a step back, we can highlight that the individual simplices composing the HexSSPs---the grid cell-like representations---are, in combination, capable of approximating any radial basis function as their kernel while demonstrating hexagonal, periodic kernels individually. In fact, the simplex shape provides optimal coverage of the $n$-dimensional feature space by ensuring equal contribution of all simplex points in all direction while minimizing the number of points it takes to realize this effect. In doing so, grid cell-like representations---perhaps grid cells themselves---provide an optimal primitive for comparing points represented in high-dimensional feature spaces.
\section{Conclusion}
In this brief account, we have demonstrated how neurally-plausible representations with grid cell-like phase structures are ``all you need'' to approximate any radial basis function. We have additionally characterized the kernel most typically used in HexSSP research, the integrated hypergeometric kernel, as well two possible alternatives: the Guassian kernel and the jinc kernel. We hope this work will motivate increased consideration into how neurally-plausible representations are constructed across a variety of tasks, including high-dimensional optimization and cognitive modelling.

\section{Acknowledgements}
The author would like to thank Chris Eliasmith for numerous discussions on Spatial Semantic Pointers---all of which helped to guide this line of research.


\printbibliography

@inproceedings{komer_neural_2019,
    title = {A neural representation of continuous space using fractional binding},
    volume = {41},
    url = {https://escholarship.org/uc/item/3zz346g1},
    booktitle = {Proceedings of the {Annual} {Meeting} of the {Cognitive} {Science} {Society}},
    author = {Komer, Brent and Stewart, Terrence C. and Voelker, Aaron R. and Eliasmith, Chris},
    year = {2019},
    pages = {2038--2043},
}

@inproceedings{rahimi_random_2007,
    title = {Random {Features} for {Large}-{Scale} {Kernel} {Machines}},
    volume = {20},
    url = {https://proceedings.neurips.cc/paper_files/paper/2007/file/013a006f03dbc5392effeb8f18fda755-Paper.pdf},
    booktitle = {Advances in {Neural} {Information} {Processing} {Systems}},
    publisher = {Curran Associates, Inc.},
    author = {Rahimi, Ali and Recht, Benjamin},
    editor = {Platt, J. and Koller, D. and Singer, Y. and Roweis, S.},
    year = {2007},
}

@inproceedings{dumont_accurate_2020,
    title = {Accurate representation for spatial cognition using grid cells},
    volume = {42},
    booktitle = {Proceedings of the {Annual} {Meeting} of the {Cognitive} {Science} {Society}},
    author = {Dumont, Nicole Sandra-Yaffa and Eliasmith, Chris},
    year = {2020},
    pages = {2367--2373},
    url = {https://escholarship.org/uc/item/8720b88v},
}

@article{dumont_exploiting_2023,
    title = {Exploiting semantic information in a spiking neural {SLAM} system},
    volume = {17},
    issn = {1662-453X},
    url = {https://www.frontiersin.org/articles/10.3389/fnins.2023.1190515/full},
    doi = {10.3389/fnins.2023.1190515},
    urldate = {2026-01-18},
    journal = {Frontiers in Neuroscience},
    author = {Dumont, Nicole Sandra-Yaffa and Furlong, P. Michael and Orchard, Jeff and Eliasmith, Chris},
    month = jul,
    year = {2023},
    pages = {1190515},
}

@book{bochner_lectures_1959,
    series = {Annals of {Mathematics} {Studies}},
    title = {Lectures on {Fourier} {Integrals}},
    publisher = {Princeton University Press},
    author = {Bochner, Solomon},
    month = sep,
    year = {1959},
}

@misc{voelker_short_2020,
    title = {A short letter on the dot product between rotated {Fourier} transforms},
    copyright = {arXiv.org perpetual, non-exclusive license},
    url = {https://arxiv.org/abs/2007.13462},
    doi = {10.48550/ARXIV.2007.13462},
    urldate = {2026-05-06},
    publisher = {arXiv},
    author = {Voelker, Aaron R.},
    year = {2020},
    note = {Version Number: 1},
}

@article{stockel_assorted_2020,
    title = {Assorted {Notes} on {Radial} {Basis} {Functions}},
    url = {https://www.researchgate.net/doi/10.13140/RG.2.2.27177.62563/1},
    doi = {10.13140/RG.2.2.27177.62563/1},
    language = {en},
    urldate = {2026-05-07},
    publisher = {University of Waterloo},
    author = {Stöckel, Andreas},
    year = {2020},
    note = {Version Number: 1},
}

\end{document}